\documentclass[sigconf,screen,nonacm,balance=false]{acmart}

\usepackage{graphicx}
\usepackage{booktabs}
\usepackage{tabularx}
\usepackage{colortbl}
\usepackage{multirow}
\usepackage{amsmath}
\usepackage{xspace}
\usepackage{enumitem}
\usepackage{url}
\usepackage[most]{tcolorbox}
\usepackage{placeins}

\setcopyright{none}
\newcommand{\skillmd}{\texttt{SKILL.md}\xspace}

\newtcolorbox{findingbox}[1]{colback=black!10,colframe=black,
  boxrule=0.6pt,arc=1.5pt,boxsep=0pt,
  left=6pt,right=6pt,top=5pt,bottom=5pt,
  before skip=7pt,after skip=7pt,
  before upper={\textbf{#1:}\enspace}}

\title{An Empirical Study of Agent Skills' Downstream Utility}
\author{Yu Cheng}
\affiliation{%
  \institution{Zhejiang University}
  \city{Hangzhou}
  \country{China}}
\email{yucheng1127@zju.edu.cn}

\author{Dehai Zhao}
\affiliation{%
  \institution{Zhejiang University}
  \city{Hangzhou}
  \country{China}}
\email{dehai.zhao@zju.edu.cn}

\author{Zhongxin Liu}
\affiliation{%
  \institution{Zhejiang University}
  \city{Hangzhou}
  \country{China}}
\email{liu_zx@zju.edu.cn}

\author{Qing Huang}
\affiliation{%
  \institution{Jiangxi Normal University}
  \city{Nanchang}
  \country{China}}
\email{qh@jxnu.edu.cn}

\author{Zhenchang Xing}
\affiliation{%
  \institution{CSIRO's Data61}
  \country{Australia}}
\email{zhenchang.xing@data61.csiro.au}

\author{Xiaoxue Ren}
\authornote{Corresponding author.}
\affiliation{%
  \institution{Zhejiang University}
  \city{Hangzhou}
  \country{China}}
\email{xxren@zju.edu.cn}

\hypersetup{
  pdfauthor={Yu Cheng, Dehai Zhao, Zhongxin Liu, Qing Huang, Zhenchang Xing, Xiaoxue Ren},
  pdfkeywords={LLM agents, agent harnesses, skill utility, skill organization}
}

\begin{document}
\begin{abstract}
Agent Skills package procedural guidance and resources for reuse, but a
relevant Skill does not necessarily improve task performance. Existing studies
characterize Skill content and evaluate downstream performance, yet provide
limited explanations of how utility depends on content, execution
configuration, and multi-Skill organization. We conduct an empirical study on
87 SkillsBench tasks, defining downstream utility as the pass-rate difference
from No-Skill on the same tasks under the same model--harness configuration.
We compare the same Skills across nine configurations, then examine
alternative published Skills and organizations of fixed Skill sets under
three selected configurations. We retrieve marketplace candidates from a
curated corpus of 37,596 Skills. LLM-assisted analysis of content, execution
traces, and final artifacts, followed by author review, relates provided
support to actual use and task outcomes. The same Skills help some
configurations and hurt others on 36.78\% of tasks, with trajectories showing
that recommended procedures can become an execution burden. Relevance
rankings overlook more useful candidates. Within the evaluated candidate
sets, reranking by support for required operations raises first-choice pass
rates by 4.35--5.80 percentage points across the three configurations. We
derive 17 authoring practices linking executable procedures to recovery,
preservation of task requirements, and checks on final artifacts. Stage Plan
and Dependency DAG outperform use order alone, with DAG's additional benefits
concentrated in tasks supplied with five or six Skills. These findings guide
developers to assess usable operation support, allow procedure adaptation
while preserving task requirements, and make artifact dependencies explicit
when organizing Skills.

\end{abstract}
\ccsdesc[500]{Software and its engineering~Empirical software validation}

\keywords{LLM agents, agent harnesses, skill utility, skill organization}
\maketitle

\section{Introduction}
\label{sec:introduction}

Agent Skills package procedural guidance, scripts, and reference materials as
reusable resources that LLM-based agents load during task execution. A Skill
typically describes procedures and usage requirements in a \texttt{SKILL.md}
file and may bundle supporting resources~\cite{anthropic2025skillspec}.
These packages can capture domain workflows, project-specific rules, or review
checklists, giving agents concrete guidance for carrying out
tasks~\cite{wu2026adaptation}. Developers can version, share, and adapt the
instruction files and bundled resources for reuse across
projects~\cite{hong2026anatomy,wu2026adaptation}. Public marketplaces distribute
tens of thousands of Skills, allowing developers to find relevant Skills and
supply agents with third-party procedures instead of having agents rediscover
them from
scratch~\cite{liu2026skillswild,ling2026skills,liang2026skillnet}.

Supplying a relevant Skill, however, does not reliably help. Prior evaluations
report that Skills can raise task pass rates, have no effect, or even lower
them~\cite{huang2026generated,li2026skillsbench,han2026sweskills,huang2026consumption}. We define the
downstream utility of a Skill condition as the difference in task pass
rate between that condition and the No-Skill condition, evaluated on the same
tasks under the same model--harness configuration. For developers reusing
published Skills, task relevance alone is insufficient to judge their utility~\cite{su2026sra}.
Skills with similar purposes can differ in the operations they cover, the
resources they provide, and the conditions their procedures
require~\cite{wu2026adaptation}. Their use also depends on how the target
model--harness configuration carries out those procedures and how intermediate
results are passed between Skills~\cite{han2026skill,chen2026skvm,li2026agentskillos,xia2026grasp}.
Without understanding how these differences affect task completion, developers
have limited guidance for selecting, adapting, or coordinating relevant Skills.

Existing studies examine Agent Skills primarily through content
analysis and downstream performance evaluation. Content studies analyze
semantic components, authoring practices, and downstream
modifications~\cite{hong2026anatomy,wu2026adaptation,liang2026ssl}. They identify current
and recommended content, but leave unclear whether these practices improve
task completion. Execution benchmarks compare pass rates
with and without Skills~\cite{li2026skillsbench,han2026sweskills}. Pass rates
indicate performance changes, but do not show which instructions agents
followed, which resources they used, or
where they encountered difficulties. In short, \emph{prior work characterizes
Skill content and measures task performance, but does not sufficiently explain
how downstream utility depends on Skill content, the model--harness
configuration, and the organization of multiple Skills.}

Explaining Skill utility through comparisons of published Skills and analysis
of their actual use involves two challenges. \emph{\textbf{Challenge-1: Useful authoring
practices must be grounded in the content and execution of real-world Skills.}}
This requires comparing relevant alternatives on the same task and evaluating
the same Skills across execution configurations and organization strategies.
\emph{\textbf{Challenge-2: Differences in Skill content may not be reflected
in what agents actually do.}} Agents may skip instructions, adapt procedures,
or supply missing operations, so execution traces and final artifacts are
needed to identify the guidance used and execution problems.

To address these challenges, we conduct an empirical study combining
real-world Skill comparisons with LLM-assisted analysis. We collect Skills
from claude-plugins.dev~\cite{claudepluginsRegistry} and
ClawHub~\cite{clawhubRegistry} and filter them for content completeness,
adoption, language, and duplication, retaining 37,596
Skills~\cite{hong2026anatomy}. From this corpus,
we retrieve relevant alternatives for tasks supplied with a single Skill and
compare their execution using each candidate's original instructions and
resources. We
then use an LLM to extract and compare task and Skill operations and identify
patterns in Skill content, execution trajectories, and final
artifacts~\cite{chen2026skvm,liang2026ssl,wu2026adaptation}. Two
authors review these patterns against the source evidence, examine alternative
explanations, and derive authoring practices from the
comparisons~\cite{runeson2009casestudy,cruzes2011thematic}. Our study
uses 87 SkillsBench tasks and their 232 supplied Skills, with three LLMs and
three agent harnesses, and addresses the following
research questions~\cite{li2026skillsbench}.

\textbf{RQ1.} How does the downstream utility of the same Skills vary across
model--harness configurations?
Aggregate gains do not show whether a Skill's benefit transfers across
configurations. We cross three models with three harnesses and compare
execution with and without the benchmark-supplied
Skills in all nine configurations, at both the overall and the task level. We
also use the results to select one configuration per model for RQ2 and RQ3.

\textbf{RQ2.} How does downstream utility differ among Skills targeting the
same task capability?
Relevance-based retrieval supplies candidates, but whether relevance predicts
utility is untested. We compare the benchmark-supplied Skill with the top five
relevance-ranked marketplace candidates on the 23 single-Skill tasks to avoid
confounding from multi-Skill coordination. We
decompose task requirements and Skill procedures into \emph{atomic
operations}---basic actions on concrete objects with identifiable
outcomes---and rerank candidates by their support for these operations.
Finally, comparisons of candidate content, trajectories, and deliverables
within candidate groups yield authoring practices grounded in observed utility
differences.

\textbf{RQ3.} How do Skill organization strategies affect downstream utility
in multi-Skill tasks?
To isolate the effect of organization, on the 41 tasks supplied with at least
three Skills we keep the Skill set and resources fixed and compare execution
without organization instructions (Flat) with three schemes that add
progressively richer information: a use order (Sequence), stage objectives
(Stage Plan), and artifact dependencies (Dependency DAG). Trajectories show how
intermediate results are passed, checked, and revised across Skills.

Our analysis exposes actionable findings for adapting, selecting, authoring,
and organizing Skills. \textbf{For utility across configurations,} Skills
help when their procedures match what the target model--harness configuration
can execute and adapt (Finding~1). The same
Skills yield gains under some configurations and losses under others on
36.78\% of tasks. Trajectories show that recommended procedures can become an
execution burden when agents struggle to implement or adapt them.
\textbf{For utility differences among Skills,} task relevance can mask gaps
in the operations a Skill supports (Finding~2a).
Lower-ranked candidates outperform the first choice in 23.19\% of
task--configuration groups.
Assessing operation support improves selection among relevant candidates
(Finding~2b). Reranking the same candidates raises
first-choice pass rates by 4.35--5.80 percentage points across three
configurations.
Skill authoring for utility links executable procedures to recovery and checks
on the final artifact (Finding~2c). From content and
execution comparisons, we summarize 17 authoring practices, including
preserving justified acceptance criteria when changing methods and specifying
which changes require rechecking final artifacts.
\textbf{For the effects of multi-Skill organization,} Stage Plan and
Dependency DAG yield greater downstream utility than Sequence across all
three configurations (Finding~3). DAG's additional
gains over Stage Plan concentrate in tasks supplied with five or six Skills,
reaching 3.03--12.12 percentage points. Trajectories show that intermediate
artifacts support source checking and propagation of corrections.

With the findings above, we can connect practices in real-world Skills to
downstream utility and distinguish the guidance Skills provide from what
agents actually use. These results offer developers concrete criteria for
selecting Skills with usable operation support, adapting procedures while
preserving task requirements, and coordinating Skills through explicit
artifact dependencies. There is a need for tools that identify unsupported
operations, assess suitability for target model--harness configurations, and
track which downstream operations and checks are affected by artifact
changes. Finally, there are opportunities for researchers to investigate
representations and evaluation methods that connect Skill content, execution
behavior, and task outcomes. The identified authoring practices provide
concrete starting points for testing which content revisions improve utility
and whether their benefits transfer across tasks and configurations.

The main contributions of this study are as follows.
\begin{itemize}[leftmargin=*,nosep]
  \item We address the gap between Skill content analysis and downstream
  performance evaluation by jointly examining published Skills, execution
  traces, and final artifacts. This analysis distinguishes supplied support
  from agents' actual use and relates these differences to task outcomes.
  \item We systematically analyze how downstream utility varies across
  model--harness configurations, relevant Skill alternatives, and organizations
  of the same Skill set. Within-task comparisons reveal the limits of
  relevance-based selection and quantify the benefits of operation-support
  reranking and organization through stage objectives and artifact dependencies.
  \item We derive a five-stage framework of 17 authoring practices grounded in
  the content and execution of real-world Skills, addressing the need for
  utility-informed authoring guidance. We discuss their implications for Skill
  development, selection, coordination tools, and future empirical evaluation.
\end{itemize}


\section{Background and Related Work}
\label{sec:background}

\subsection{LLM Agents and Agent Skills}
Agent Skills are read and used by LLM agents during task execution. Their utility depends on how models interpret the instructions and on the tools and control mechanisms provided by the execution framework. LLMs such as GPT, DeepSeek, and Qwen generate action decisions from task instructions and environmental feedback~\cite{schick2023toolformer,qin2024toolllm}, following the interleaving of reasoning and action exemplified by ReAct~\cite{yao2023react}. An \emph{agent harness} manages context, tool calls, and execution control. SWE-agent and OpenHands investigate agent interface design and general-purpose software execution platforms, respectively~\cite{yang2024sweagent,wang2025openhands}. For example, Codex and OpenCode support coding tasks through file operations and command execution~\cite{openai2026codexcli,opencode2026docs}. OpenClaw connects models to external tools and services for general-purpose tool use and automation~\cite{openclawDocs}. Skill-Use shows that the same model can exhibit different Skill-use behavior across harnesses~\cite{han2026skill}. Skill utility thus depends on the model--harness pairing, which we call a \emph{model--harness configuration}.

Agent Skills provide reusable procedural knowledge through instructions and supporting resources. Under the Agent Skills specification, each Skill contains a required \skillmd file comprising YAML metadata and a Markdown body, optionally accompanied by scripts, references, and assets. Progressive disclosure first exposes the name and description for discovery, loads the body upon activation, and accesses supporting resources as needed~\cite{anthropic2025skillspec}. This structure separates information for Skill selection, instructions for task execution, and resources available during use. Public skill marketplaces enable developers to distribute and reuse these materials~\cite{liu2026skillswild,ling2026skills}. SkillNet further provides infrastructure for Skill creation, evaluation, and relational organization~\cite{liang2026skillnet}. Procedural knowledge reuse also appears in experience-based learning~\cite{shinn2023reflexion,xia2026skillrl}, including natural-language insights in ExpeL~\cite{zhao2024expel}, executable programs in Voyager and ASI~\cite{wang2024voyager,wang2025programmatic}, and task workflows in Agent Workflow Memory~\cite{wang2025awm}. Our focus is on published Skills with procedural instructions and supporting resources, specifically their downstream utility when supplied to agents.

\subsection{Skill Retrieval and Organization}
Skill retrieval research examines how to select relevant support for a task from a candidate library. Related tool reuse research has established key challenges: general-purpose retrievers struggle with specialized tools~\cite{shi2025toolret}, and structured documentation improves matching~\cite{lu2026tooldocs}. 
Skill-specific approaches push further. SkillRet introduces a large-scale retrieval benchmark and finds that Skill-specific training outperforms general-purpose retrievers~\cite{cho2026skillret}. SkillRouter shows that names and descriptions alone omit useful routing information and reports task-performance gains from its full-text retrieval and reranking pipeline~\cite{zheng2026skillrouter}. Yet SRA finds that retrieving a relevant Skill does not ensure appropriate decisions about loading and use~\cite{su2026sra}, highlighting a gap between retrieval quality and downstream performance. We compare observed utility within a fixed, relevance-filtered candidate set to assess whether operation support helps candidate selection.

Structured representations and compilation further characterize the operations and execution conditions required by Skills. SSL separates scheduling information, execution structure, and evidence of actions and resource use, applying this representation to Skill discovery and risk assessment~\cite{liang2026ssl}. SkVM decomposes Skill requirements into primitive capabilities, assesses support from model--harness configurations, and adapts execution through compilation and runtime mechanisms~\cite{chen2026skvm}. SkCC uses an intermediate representation to separate Skill semantics from platform-specific formats for deployment across frameworks~\cite{ouyang2026skcc}. Such representations provide a basis for extracting comparable operations and constraints from natural-language content. We adopt this operation decomposition to analyze support for task requirements, while using the original Skill content and resources for execution.

Multi-Skill organization research examines how selected capabilities are connected~\cite{zeng2026groupskills}. Graph of Skills incorporates dependencies into retrieval to return Skill sets within a context budget~\cite{li2026graphskills}. SkillDAG uses typed edges to represent dependencies, conflicts, and other relationships, allowing agents to update the graph using execution evidence~\cite{bai2026skilldag}. GraSP converts Skill sets into graphs with precondition and effect relationships, combining node verification with local repair during execution~\cite{xia2026grasp}. AgentSkillOS organizes discovery through a capability tree and coordinates execution through a DAG, reporting higher artifact quality than flat invocation with the same Skill set~\cite{li2026agentskillos}. These findings provide evidence for explicit organization. We compare the utility of four organization schemes while holding the original Skills and resources fixed. These schemes provide no additional organization instructions, a use sequence, stage objectives, or artifact dependencies. Execution trajectories let us trace how agents pass intermediate results between Skills, check them, and revise them under each organization scheme.

\begin{figure*}[t]
  \centering
  \includegraphics[width=\linewidth]{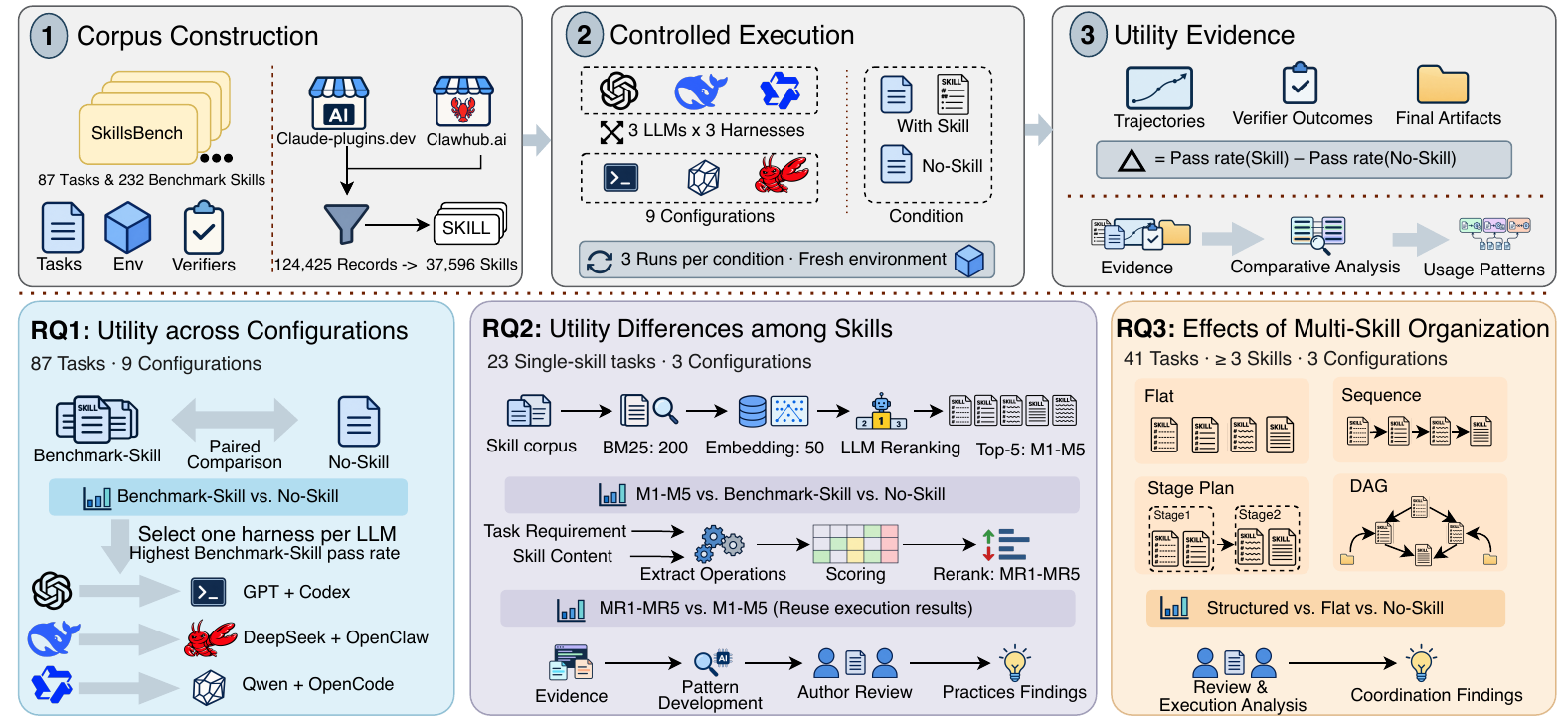}
  \caption{Overview of the empirical study on Agent Skill utility.}
  \Description{Overview of the empirical study. A shared Skill corpus, task benchmark, and experimental setup support three research questions on model--harness configurations, alternative Skills for the same task, and multi-Skill organization. Content and trajectory analysis link observed utility differences to Skill authoring practices.}
  \label{fig:methodology}
\end{figure*}

\subsection{Empirical Studies of Agent Skills}
Existing empirical studies examine Agent Skills primarily through content analysis and downstream performance evaluation. Content studies characterize what Skills contain and how developers revise them. Hong et al.\ analyze the semantic components of \skillmd, synthesize authoring practices through a multivocal literature review, and define violations as Skill smells to study their prevalence and evolution~\cite{hong2026anatomy}. Wu et al.\ examine downstream modifications, identifying adaptation patterns and relationships among changes to procedures, resources, and local environments~\cite{wu2026adaptation}. These studies establish content requirements and recurring adaptation needs, but do not test whether alternative published Skills following different practices improve completion of the same tasks. Our authoring analysis grounds practices in comparisons of published Skill content, the procedures agents actually follow, and the outcomes they achieve on the same tasks.

Downstream evaluations measure whether Skills improve performance and whether agents use their guidance. SkillsBench compares execution without Skills, with curated Skills, and with model-generated Skills across domains using programmatic verification~\cite{li2026skillsbench}. SWE-Skills-Bench evaluates Skills in software repositories, reporting limited gains and regressions involving conflicts with project context~\cite{han2026sweskills}. Skill-Use examines triggering, procedural compliance, and operational boundaries across harnesses~\cite{han2026skill}. These studies establish outcome differences and obstacles to Skill use. We compare relevant alternatives for the same task under the same configuration, tracing which procedures and resources agents use, skip, or supplement. This distinguishes support described in Skills from support used in task execution.

\begin{table}[t]
  \caption{Skill corpus construction.}
  \label{tab:skill-corpus-pipeline}
  \centering
  \small
  \setlength{\tabcolsep}{2pt}
  \begin{tabular}{@{}lccccc@{}}
    \toprule
    Stage & Collection & Completeness & Popularity & Language & Deduplication \\
    \midrule
    Records & 124,425 & 113,972 & 65,048 & 40,412 & 37,596 \\
    \bottomrule
  \end{tabular}
\end{table}

\begin{table}[t]
  \caption{Distribution of tasks by supplied Skill count.}
  \label{tab:task-skill-distribution}
  \centering
  \begin{tabular}{lrrrrrrr}
    \toprule
    Skills per task & 1 & 2 & 3 & 4 & 5 & 6 & 7 \\
    Tasks & 23 & 23 & 20 & 9 & 6 & 5 & 1 \\
    \bottomrule
  \end{tabular}
\end{table}

Some studies also connect content changes to downstream outcomes when improving Skill generation and management. Huang et al.\ compare experience generation, Skill extraction, and consumption, deriving content evaluation criteria and a meta-skill for subsequent extraction~\cite{huang2026consumption}. ASSAY estimates Skill contributions to restructure libraries and select Skills for individual tasks~\cite{wang2026assay}. SkillReducer compresses and restructures content to reduce token costs~\cite{gao2026skillreducer}, while SkillMOO optimizes Skill bundles for pass rate and inference cost and analyzes effective edits~\cite{gong2026skillmoo}. We compare relevant published Skills using their original instructions and resources, and examine configuration and organization effects while holding the supplied Skills fixed. Joint analysis of content, trajectories, and final artifacts relates these utility differences to observed Skill use, providing evidence for authoring, selection, and coordination decisions.

\section{Methodology}
\label{sec:methodology}
We combine controlled execution with content and trajectory analysis to link task requirements, Skill content, observed use, and task outcomes,
examining how Skill content and organization relate to downstream utility. As Figure~\ref{fig:methodology} shows,
RQ1 examines the utility of the same Skills across model--harness configurations and selects configurations for subsequent analyses;
RQ2 compares alternative Skills for the same task to derive content design insights;
and RQ3 compares organizations of the same Skill set to examine cross-Skill coordination.

\subsection{Corpus Construction}
\label{subsec:corpus-construction}

\noindent\textbf{\textit{Skill Curation.}}
To study third-party Skill reuse, we collect candidate Skills from two public platforms serving different agent ecosystems: \url{https://claude-plugins.dev},
a community-maintained platform for Claude Code plugins and Agent Skills~\cite{claudepluginsRegistry},
and ClawHub (\url{https://clawhub.ai}), OpenClaw's public Skill registry~\cite{clawhubRegistry}.
The snapshots we collect on August 20, 2026 contain 47,645 and 76,780 records, respectively, for a total of 124,425 records.
Following prior work~\cite{hong2026anatomy}, we filter the corpus for content completeness, adoption, language, and duplication, as summarized in Table~\ref{tab:skill-corpus-pipeline}.
First, records must have nonempty names, descriptions, and bodies, since retrieval and subsequent content analysis depend on all three fields.
Second, we keep records with at least 100 downloads as evidence of actual use.
Third, names and descriptions must be English, and non-Latin letters must account for at most 3\% of body letters, which limits language variation.
Finally, we identify duplicates by hashing normalized Skill bodies with SHA-256.
For normalization, we remove frontmatter and Markdown formatting, lowercase the body, and remove nonalphanumeric characters.
We retain one representative per duplicate group, prioritizing download count, then star count, then source identifier.
This leaves 37,596 Skills for candidate retrieval.

\noindent\textbf{\textit{Task Benchmark.}} We select SkillsBench as the task benchmark for studying Skill utility.
It provides tasks with associated Skills, executable environments, and programmatic verifiers,
enabling utility assessment through task execution and outcome verification~\cite{li2026skillsbench}.
We use 87 tasks spanning eight domains, collected as of September 20, 2026~\cite{benchflow2026skillsbenchtasks}.
Table~\ref{tab:task-skill-distribution} summarizes their distribution by the number of supplied Skills.
These tasks include 232 benchmark-supplied Skills, which we refer to as \emph{Benchmark Skills}.
We select task subsets according to each RQ's comparison objective.
RQ1 uses all 87 tasks to examine the utility of the same Skills across model--harness configurations.
RQ2 uses the 23 tasks supplied with a single Skill to compare how alternative Skills support the same task.
RQ3 uses the 41 tasks supplied with at least three Skills to compare how ordering, stage grouping,
and dependencies for the same Skill set affect downstream utility.

\subsection{Controlled Execution}
\label{subsec:experimental-setup}

\noindent\textbf{\textit{LLMs and Agent Harnesses.}} We select a proprietary LLM and two open-weight LLMs with different architectures and parameter scales~\cite{huang2026consumption,su2026sra}. GPT-5.5 is a proprietary model for coding and tool-based tasks~\cite{openai2026gpt55}. DeepSeek-V4-Flash is an open-weight mixture-of-experts (MoE) model with 284B total and 13B activated parameters~\cite{deepseek2026v4flash}. Qwen3.5-27B is an open-weight 27B dense model~\cite{qwen2026qwen35_27b}. We select Codex and OpenCode for coding-oriented execution~\cite{openai2026codexcli,opencode2026docs}, and OpenClaw for general-purpose tool use and automation~\cite{openclawDocs}. Each harness retains its native interfaces and tools, as well as the mechanisms it uses to control task execution.

\noindent\textbf{\textit{Task Execution.}} Skill conditions provide the corresponding Skills, and the task instruction explicitly asks agents to consult and use them.
The \emph{No-Skill} condition provides neither Skills nor the corresponding consultation and use request.
We run each task under each condition three times per configuration, using a fresh environment for every run.

\subsection{Utility Evidence}
\label{subsec:utility-evidence}

\noindent\textbf{\textit{Utility Measurement.}} After each execution, the task's original verifier assesses the final state: runs that pass verification are recorded as successful, and all others as unsuccessful.
We measure task performance by pass rate and downstream utility by the pass-rate change relative to No-Skill~\cite{huang2026consumption}.
As in SkillsBench~\cite{li2026skillsbench}, repeated outcomes are first averaged within each task and then averaged equally across tasks.
Let $p_{t,c,x}$ denote the fraction of three runs that pass for task $t$ under configuration $c$ and condition $x$. For task set $T$, the pass rate is
\begin{equation}
  P_c(x;T)=\frac{1}{|T|}\sum_{t\in T}p_{t,c,x}.
  \label{eq:pass-rate}
\end{equation}
On the same task set and model--harness configuration, we define the downstream utility of a Skill condition relative to No-Skill as
\begin{equation}
  \Delta_c(x;T)=P_c(x;T)-P_c(\text{No-Skill};T).
  \label{eq:skill-utility}
\end{equation}
The sign of $\Delta_c(x;T)$ indicates a pass-rate gain or loss relative to No-Skill.

\noindent\textbf{\textit{Execution Evidence.}} Building on outcome evaluation, we examine execution trajectories and final artifacts to compare Skill use in actual operations and relate it to task outcomes across conditions. Subsequent RQ sections describe case selection and analysis.

\subsection{RQ1: Skill Utility across Model--Harness Configurations}
\label{subsec:rq1-method}

To examine how the utility of the same Skills varies across model--harness configurations, we cross three LLMs with three agent harnesses and evaluate all nine configurations on the 87 tasks. Each configuration includes Benchmark-Skill and No-Skill conditions; for a given task, the Benchmark-Skill condition provides the same benchmark-supplied Skills and resources across configurations. Execution follows Section~\ref{subsec:experimental-setup}, with evaluation defined in Section~\ref{subsec:utility-evidence}.

For each configuration, we compute overall pass rates under both conditions and the gain from Benchmark-Skill relative to No-Skill. We compare gains across harnesses for the same model and across models under the same harness, and compare model pass-rate gaps under the two conditions within each harness. At the task level, we distinguish four groups based on gains and losses across the nine configurations, comprising gains without losses, losses without gains, both gains and losses, and no changes. For tasks showing both gains and losses, we compare case trajectories across configurations to examine Skill resource use and the handling of task constraints.

Based on the overall RQ1 results, we select the harness with the highest Benchmark-Skill pass rate for each model, yielding three configurations for comparing the utility of Skill content and organization in RQ2 and RQ3. This retains all three models while limiting the experiment scale. We compare candidates and organization schemes within each configuration.

\subsection{RQ2: Utility Differences among Skills Targeting the Same Capability}
\label{subsec:rq2-method}

RQ2 uses 23 tasks, each supplied with one Benchmark Skill, so that alternative Skills for the same task can be compared and their differences examined through content and execution evidence. This focuses the comparison on individual Skill support and reduces variation from multi-Skill coordination. Under the three configurations selected in RQ1, we compare candidate utility, examine operation information for selection, and trace content use during execution.

\subsubsection{Alternative Skill Selection}
\label{subsubsec:rq2-candidate-selection}

Comparing downstream utility across Skills for the same task first requires alternatives relevant to its requirements, which we select from the marketplace corpus in Section~\ref{subsec:corpus-construction}. We use a retrieve-then-rerank framework~\cite{cho2026skillret,su2026sra}, with the original task instruction as the query. We exclude records sharing the normalized body or source URL of the task's benchmark-supplied Skill, then retrieve 200 candidates with BM25~\cite{robertson2009bm25}. We encode the task instruction and each candidate's name and description using \texttt{text-embedding-3-large} and retain the 50 candidates with the highest cosine similarity.

Beyond names and descriptions, the \skillmd body provides specific procedures and applicability conditions~\cite{zheng2026skillrouter}. We use Claude-opus-4.6 to rank these 50 candidates using the task instruction and their names, descriptions, and bodies~\cite{sun2023rankgpt}. We call the top five Skills \emph{marketplace candidates} and denote them M1--M5 in relevance order. Each rank refers to a candidate for that task, rather than a single Skill shared across all tasks in the benchmark.

Each task has seven conditions comprising Benchmark-Skill, five marketplace candidates, and No-Skill. Under the three configurations selected in RQ1, each candidate condition provides one Skill with its original content and supporting resources. Execution and evaluation follow Sections~\ref{subsec:experimental-setup} and~\ref{subsec:utility-evidence}, respectively. We compute each condition's overall pass rate and each Skill condition's downstream utility relative to No-Skill. Marketplace candidates are aggregated across tasks by relevance rank M1--M5 to examine how relevance rankings relate to observed utility. We also compare candidate pass rates within each task and configuration, providing contrasts for atomic-operation reranking and subsequent content and execution analysis.

\subsubsection{Atomic-operation Reranking}
\label{subsubsec:rq2-operation-support}

The candidate utility comparison shows that Skills with higher relevance ranks do not necessarily yield higher downstream utility (Section~\ref{subsec:rq2-results}). We observe that candidates for the same task differ in operation coverage, inputs and outputs, and applicability conditions. We therefore rerank candidates by examining how their procedures support required task operations and which steps the agent still needs to supply.

We decompose task requirements and Skill procedures into \emph{atomic operations} for item-by-item comparison, defining them as basic operation units explicitly described in the source that act on concrete objects and have identifiable outcomes. We use an LLM to extract actions, objects, inputs or parameters, and constraints from task instructions~\cite{chen2026skvm}. For Skills, we extract procedures or interfaces from \skillmd and available supporting files, recording objects, inputs and outputs, applicability conditions, and resources, along with source locations and excerpts~\cite{chen2026skvm,liang2026ssl}. These representations support assessment, while execution uses the original Skill content. We use Claude-opus-4.6 to extract both task and Skill operations.

For each task operation, we assess whether the candidate's procedures meet the object, input, output, and constraint requirements, distinguishing support that requires only task-parameter binding from support that requires additional substantive logic. Multiple Skill operations may jointly meet a requirement, but their connections must have source evidence, and missing steps cannot be supplied to establish full support. We use Claude-opus-4.6 for LLM-based support assessment~\cite{ahmed2025annotation,wang2025llmjudge}, providing the task's atomic operations, the candidate Skill's operation descriptions, and the corresponding source excerpts. The model applies these criteria to each task operation and returns one of the following support labels in JSON format.

\begin{itemize}[leftmargin=*,itemsep=2pt,topsep=4pt,parsep=0pt]
  \item \textbf{Full ($1$).} The required operation can be performed through task-parameter binding alone, with no identified conflict with the task's explicit constraints.
  \item \textbf{Partial ($0.5$).} Relevant support is provided without an identified constraint conflict, but additional substantive logic or decisions are required.
  \item \textbf{Absent ($0$).} No evidence of support for the operation is found.
  \item \textbf{Conflicting ($-1$).} The provided procedure violates an explicit task constraint, and the Skill provides no alternative compatible with that constraint.
\end{itemize}

To rank candidates, we assign heuristic scores of 1, 0.5, 0, and $-1$ to Full, Partial, Absent, and Conflicting judgments, respectively. These values distinguish full support, support requiring additional logic, missing support, and explicit conflicts. We average scores equally across task operations. Let $\mathcal{O}_t$ denote the atomic operation set for task $t$, and let $N_{\mathrm{F}}$, $N_{\mathrm{P}}$, and $N_{\mathrm{C}}$ count the Full, Partial, and Conflicting judgments for candidate Skill $s$. The score is
\begin{equation}
  \operatorname{Match}(t,s) =
  \frac{N_{\mathrm{F}}+0.5N_{\mathrm{P}}-N_{\mathrm{C}}}{|\mathcal{O}_t|}.
  \label{eq:atomic-operation-match}
\end{equation}
We rerank the five marketplace candidates by descending score as MR1--MR5 and reuse execution results to aggregate pass rates by rank. Within each configuration, reranking gain is the mean task-level pass-rate difference between MR1 and M1 (Section~\ref{subsec:utility-evidence}).

\subsubsection{Utility-Grounded Authoring Practices}
\label{subsubsec:rq2-authoring-practices}

Our unit of analysis is a candidate comparison group for the same task and configuration, used to identify authoring requirements associated with downstream utility. We combine pass-rate differences with Skill content, execution trajectories, and final deliverables to examine how content requirements are adopted and the corresponding task outcomes. Drawing on studies of prerequisites and adaptation needs~\cite{wu2026adaptation}, procedural guidance and verification~\cite{hong2026anatomy}, and program refinement through execution feedback and self-verification~\cite{wang2024voyager,sun2023adaplanner}, we organize observations and practices into five stages of Skill use, namely Preparation, Adaptation, Implementation, Verification, and Correction. Our analysis followed three steps comprising case comparison, LLM induction, and author review.

We selected the initial sample from execution outcomes. The 1,449 RQ2 runs form 69 comparison groups, each containing three runs under each of seven conditions. Five groups contrast marketplace candidates that pass all three runs with those that fail all three, and another five contain this contrast between the benchmark-supplied Skill and marketplace candidates. These ten groups form the initial sample, covering nine tasks and 210 trajectories, including each group's No-Skill runs. Using the operation representations in Section~\ref{subsubsec:rq2-operation-support}, we compared the procedures, resources, and constraints provided by each candidate, checking whether requirements were present, read, or followed, and comparing execution behavior and deliverables.

Following prior work~\cite{wu2026adaptation}, we used Claude-opus-4.6 to identify candidate patterns in these comparisons. The model received task requirements, original Skill content and resources, operation representations from both sides, execution trajectories, final deliverables, and verification results. It examined relationships between content requirements, observed behavior, and outcomes across the five stages. For each candidate pattern, the model returned its name, description, content and execution evidence, and potential counterexamples in JSON format. Source locations, trajectory excerpts, or deliverable evidence linked observations to interpretations~\cite{runeson2009casestudy}.

Two authors checked the patterns and source evidence, distinguishing missing content requirements, requirements that are present but not followed, and support supplied by the agent during execution. They also examined alternative explanations for outcome differences. The authors merged similar patterns, clarified category boundaries, resolved disagreements through discussion, and revised or removed unsupported explanations~\cite{cruzes2011thematic}. We added cases from groups with smaller or no outcome differences to examine recurrence and scope and revisited previously analyzed cases. The two authors then derived concrete authoring requirements, specifying the operations or resources concerned, applicability conditions, and supporting evidence, and grouped them into the five-stage framework by their primary stage of use. For each practice, we record the corresponding tasks and configurations, content and execution evidence, and counterexamples. We distinguish insights based on recurring observations from those based on a single case.

\subsection{RQ3: Effects of Multi-Skill Organization on Utility}
\label{subsec:rq3-method}

We compare the downstream utility of the same Skill set under different use orders, stage groupings, and artifact dependencies. The experiments use 41 tasks supplied with at least three Benchmark Skills under the three configurations selected in RQ1. All Skill conditions retain the complete Skill set, original content, and resources, with instructions organizing their use.

Based on task planning~\cite{wang2023plansolve,singh2023progprompt} and dependency-based coordination~\cite{kim2024llmcompiler,li2026agentskillos,liu2024controlllm}, we compare Sequence, Stage Plan, and Dependency DAG against Flat, which provides no task-specific organization instructions. Instructions may specify subsets or repeated uses of Skills.

\begin{itemize}[leftmargin=*,itemsep=2pt,topsep=4pt,parsep=0pt]
  \item \textbf{Flat.} Original Skills are provided with a general instruction to consult and use them, without task-specific organization instructions.
  \item \textbf{Sequence.} The order in which Skills should be used is specified.
  \item \textbf{Stage Plan.} Execution is divided into ordered stages, each identifying the Skills to use and specifying the intermediate objective they should achieve.
  \item \textbf{Dependency DAG.} Each node represents one use of a Skill. Prerequisite dependencies and artifacts passed between nodes are explicitly specified, with subsequent nodes instructed to execute only when the required artifacts are available.
\end{itemize}

The authors manually constructed the experimental organization instructions using only SkillsBench task descriptions and the corresponding Skills. Another author checked whether use orders, stage objectives, and artifact dependencies were supported by the source materials~\cite{runeson2009casestudy}. The review distinguished explicitly specified relationships from author choices and resolved disagreements through evidence review and discussion.

Execution and evaluation follow Sections~\ref{subsec:experimental-setup} and~\ref{subsec:utility-evidence}, respectively. We report each condition's overall pass rate and downstream utility relative to No-Skill, measuring organization gains by the mean task-level pass-rate difference from Flat. We also group tasks by the number of supplied Skills to compare pass rates and organization gains across groups. For execution analysis, we followed the case comparison, LLM induction, and author review procedure in Section~\ref{subsubsec:rq2-authoring-practices}. We compared organization conditions for the same task and configuration, checking whether use orders were followed, stage objectives were met, and intermediate artifacts were read, used, or revisited by subsequent operations. Cases with pass-rate gains and losses informed design insights for cross-Skill coordination and their scope of applicability.

\section{Results}
\label{sec:results}
\suppressfloats[t]

\subsection{RQ1: Skill Utility across Model--Harness Configurations}
\label{subsec:rq1-results}

\begin{table}[t]
  \caption{Pass rates and gains over No-Skill across model--harness configurations. Bold values indicate the highest Benchmark-Skill pass rate and gain within each model.}
  \label{tab:rq1-pass}
  \centering\small
  \setlength{\tabcolsep}{4pt}
  \begin{tabular}{@{}lrrr@{}}
    \toprule
    Configuration & No-Skill (\%) & Benchmark-Skill (\%) & Gain (pp) \\
    \midrule
    GPT + Codex         & 42.15 & \textbf{54.02} & 11.88 \\
    GPT + OpenClaw      & 30.65 & 47.89 & 17.24 \\
    GPT + OpenCode      & 24.14 & 43.68 & \textbf{19.54} \\
    \midrule
    DeepSeek + Codex    & 34.10 & 44.44 & 10.34 \\
    DeepSeek + OpenClaw & 35.63 & \textbf{50.19} & \textbf{14.56} \\
    DeepSeek + OpenCode & 36.40 & 48.28 & 11.88 \\
    \midrule
    Qwen + Codex        & 2.30  & 6.90  & 4.60 \\
    Qwen + OpenClaw     & 3.45  & 17.24 & \textbf{13.79} \\
    Qwen + OpenCode     & 13.41 & \textbf{24.52} & 11.11 \\
    \bottomrule
  \end{tabular}
\end{table}

As Table~\ref{tab:rq1-pass} shows, Benchmark Skills improve the overall pass rate in all nine configurations, with gains of 4.60--19.54 percentage points over No-Skill. The same Skills also widen the performance gap between GPT and Qwen. Under the same harness, GPT already has a higher No-Skill pass rate and obtains a Skill gain 3.45--8.43 percentage points larger than Qwen's. Under OpenCode, GPT and Qwen gain 19.54 and 11.11 percentage points, respectively, widening their pass-rate gap from 10.73 to 19.16 percentage points. In the GPT--Qwen comparison, the model with higher No-Skill performance also obtains larger Skill gains under all three harnesses.

\begin{figure*}[t]
  \centering
  \includegraphics[width=\linewidth]{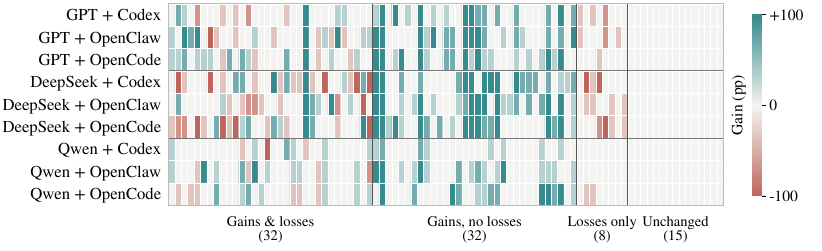}
  \caption{Task-level gains over No-Skill from the same Benchmark Skills across nine configurations. Rows and columns represent configurations and tasks, respectively. Tasks are grouped by gains and losses across configurations, with group sizes in parentheses.}
  \Description{A heatmap with nine configuration rows and 87 task columns. Teal indicates a pass-rate increase relative to No-Skill, red a decrease, and light gray no change. Thirty-two tasks show both gains and losses across configurations; 32 show gains with no losses; eight show losses with no gains; and 15 show no change in any configuration. Tasks are ordered alphabetically within each group.}
  \label{fig:rq1-task-utility}
\end{figure*}

However, the same Skills can yield gains on a task in one configuration and losses in another. Figure~\ref{fig:rq1-task-utility} shows this pattern in 32 of the 87 tasks (36.78\%), and the differences appear both across models under the same harness and across harnesses for the same model. What varies with the configuration is therefore not only the size of a Skill's gain but whether it helps at all.

Execution trajectories further show that Skill-recommended procedures can become an execution burden, with utility depending on whether the configuration can carry them out effectively. For example, \texttt{exam-\allowbreak block-\allowbreak sequencing} allows solvers, heuristics, or hybrid methods, while its supplied Skills prioritize integer programming. DeepSeek + Codex passes all three No-Skill runs but times out in all three Benchmark-Skill runs. One successful No-Skill run uses simulated annealing and local search. In contrast, one Benchmark-Skill run omits count coefficients from the integer program's objective function, then repeatedly debugs the search process after correcting the formulation, eventually timing out. GPT passes all three runs with the same Skills under the same harness. In one run, its solver process is also terminated, but it then switches to an allowed heuristic method, retains a feasible solution, and checks the final files before delivery. This contrast suggests that when a configuration struggles to carry out Skill-recommended procedures, the guidance can add execution burden and reduce task performance below No-Skill.

Following the selection rule in Section~\ref{subsec:rq1-method}, RQ2 and RQ3 use the configuration with the highest Benchmark-Skill pass rate for each model, namely GPT + Codex (54.02\%), DeepSeek + OpenClaw (50.19\%), and Qwen + OpenCode (24.52\%).

\begin{findingbox}{Finding 1}
\textbf{Skill utility depends on the fit between supplied procedures and the target model--harness configuration's ability to execute and adapt them.}
Recommended procedures can become an execution burden, even on tasks the configuration can solve without Skills.
\end{findingbox}

\subsection{RQ2: Utility Differences among Skills Targeting the Same Capability}
\label{subsec:rq2-results}
\suppressfloats[t]

We first compare the utility of alternative Skills for the same task, then examine whether operation support improves candidate selection, and finally relate Skill content and observed execution to authoring practices associated with utility differences.

\subsubsection{Utility Differences among Alternative Skills}

\begin{table}[t]
  \caption{Pass rates (\%) by original relevance rank.}
  \label{tab:rq2-pass}
  \centering\small
  \setlength{\tabcolsep}{2pt}
  \begin{tabular}{@{}lrrrrrrr@{}}
    \toprule
    Configuration & No-Skill & \shortstack{Benchmark-\\Skill} & M1 & M2 & M3 & M4 & M5 \\
    \midrule
    \shortstack[l]{GPT +\\Codex} & 36.23 & \textbf{49.28} & 42.03 & 40.58 & 43.48 & 37.68 & 37.68 \\
    \shortstack[l]{DeepSeek +\\OpenClaw} & 28.99 & \textbf{44.93} & 40.58 & 39.13 & 37.68 & 42.03 & 36.23 \\
    \shortstack[l]{Qwen +\\OpenCode} & 17.39 & 18.84 & 14.49 & \textbf{20.29} & 14.49 & 15.94 & 17.39 \\
    \bottomrule
  \end{tabular}
\end{table}

Relevant Skills for the same task do not provide equivalent downstream utility. Among the 69 comparison groups formed by 23 tasks and three configurations, 27 (39.13\%) have different pass rates across the five marketplace candidates. Table~\ref{tab:rq2-pass} aggregates candidate pass rates by original relevance rank M1--M5 and shows that candidates selected for relevance can still yield losses. Under Qwen + OpenCode, M1, M3, and M4 have lower overall pass rates than No-Skill, only M2 exceeds it, and M5 ties it. Relevance to a task does not guarantee an improvement in task performance.

Higher relevance also need not translate into larger gains. In 16 of the 69 comparison groups (23.19\%), the first-choice candidate M1 has a lower pass rate than at least one lower-ranked candidate. When aggregated by rank, the highest-performing marketplace groups are M3 (GPT + Codex), M4 (DeepSeek + OpenClaw), and M2 (Qwen + OpenCode). Overall relevance ranking misses utility differences of this kind, leaving more useful candidates below the first choice.

A candidate can match the task topic while leaving key operations for the agent to supply. The original first choice for \texttt{lab-\allowbreak unit-\allowbreak harmonization}, M1 (\texttt{medical-\allowbreak unit-\allowbreak converter}), provides medical unit conversion, but its interface requires known source and target units. The task also requires removing incomplete records, normalizing scientific notation and decimal formats, and using physiological ranges to identify values requiring conversion. Using M1 still leaves the agent to implement unit identification and additional cleaning steps, and operational gaps of this kind are not fully reflected in the overall relevance ranking. Candidate selection also needs to distinguish the support provided from the steps the agent must supply.

\begin{findingbox}{Finding 2a}
\textbf{Relevance ranking overlooks Skills with higher downstream utility.}
Highly ranked candidates can match the task topic while leaving operations for the agent to implement.
\end{findingbox}

\subsubsection{Utility Gains from Atomic-operation Reranking}

\begin{table}[t]
  \caption{Pass rates (\%) and reranking gains over M1.}
  \label{tab:rq2-atomic-pass}
  \centering\small
  \setlength{\tabcolsep}{3pt}
  \begin{tabular}{@{}lrrrrr@{}}
    \toprule
    Configuration & No-Skill & \shortstack{Benchmark-\\Skill} & M1 & MR1 & Gain (pp) \\
    \midrule
    GPT + Codex & 36.23 & 49.28 & 42.03 & 46.38 & +4.35 \\
    DeepSeek + OpenClaw & 28.99 & 44.93 & 40.58 & 44.93 & +4.35 \\
    Qwen + OpenCode & 17.39 & 18.84 & 14.49 & 20.29 & +5.80 \\
    \bottomrule
  \end{tabular}
\end{table}

Reranking by atomic operation support improves the overall utility of the first-choice Skills in all three configurations. MR1, the first choice after reranking, improves pass rates over the original first choice M1 by 4.35, 4.35, and 5.80 percentage points under GPT + Codex, DeepSeek + OpenClaw, and Qwen + OpenCode, respectively (Table~\ref{tab:rq2-atomic-pass}). Under Qwen + OpenCode, the first-choice pass rate rises from 14.49\% to 20.29\%, turning a loss relative to No-Skill into a gain. The candidate set, Skill content, and execution records remain unchanged. This support information complements overall relevance judgments in identifying Skills with higher downstream utility.

Atomic operation matching distinguishes the specific support candidates provide and the prerequisites for using it. The score in Section~\ref{subsubsec:rq2-operation-support} accounts for directly usable support, support requiring additional substantive logic, absent support, and constraint conflicts, relating candidate scores to individual task requirements. On \texttt{lab-\allowbreak unit-\allowbreak harmonization}, MR1 is the original M2 (\texttt{clinical-\allowbreak data-\allowbreak cleaner}). Whereas M1's conversion interface requires known source and target units, M2 provides unit-standardization guidance, a \texttt{drop} strategy for removing incomplete rows, operations for detecting outliers using clinical thresholds, and reference-range checks. These provisions address the required cleaning and anomaly-detection steps beyond conversion between known units. Although both candidates concern medical unit processing, they differ in the operations supported and the steps left for the agent to supply.

Execution trajectories further show how this support enters task processing. In one successful GPT + Codex run with M2, the agent reads the Skill's threshold reference file, removes incomplete rows, converts numeric formats, and processes units for each feature. It also checks output formats, missing values, and numeric ranges. This run links the candidate's reference resources to concrete processing steps, showing how operation support is used during execution. Together with the overall reranking gains, these observations suggest that Skill utility depends not only on relevance to the task but also on usable support for its required operations.

\begin{findingbox}{Finding 2b}
\textbf{Selecting Skills by their support for required operations improves downstream utility.}
Within the same candidate sets, atomic-operation reranking raises first-choice pass rates by 4.35--5.80 percentage points over relevance ranking.
\end{findingbox}

\begin{table}[t]
  \caption{\textbf{Five-stage Skill authoring framework with 17 practices.}}
  \label{tab:rq2-best-practices}
  \centering
  \footnotesize
  \setlength{\tabcolsep}{2pt}
  \renewcommand{\arraystretch}{1.05}
  \definecolor{rqTwoP}{HTML}{516B93}
  \definecolor{rqTwoA}{HTML}{498798}
  \definecolor{rqTwoI}{HTML}{447799}
  \definecolor{rqTwoV}{HTML}{927345}
  \definecolor{rqTwoC}{HTML}{9B6469}
    \begin{tcolorbox}[enhanced,frame hidden,sharp corners,boxrule=0pt,
      colback=white,colbacktitle=rqTwoP!14!white,coltitle=rqTwoP!65!black,
      borderline west={2.5pt}{0pt}{rqTwoP},
      left=5pt,right=1pt,top=2pt,bottom=2pt,boxsep=0pt,
      toptitle=3pt,bottomtitle=3pt,before skip=0pt,after skip=0pt,
      fonttitle=\footnotesize\sffamily,
      title={\textbf{PREPARATION}\enspace {\scriptsize 4 practices}}]
      \begin{tabularx}{\linewidth}{@{}>{\raggedright\arraybackslash}p{0.035\linewidth}>{\scriptsize\raggedright\arraybackslash}p{0.31\linewidth}>{\raggedright\arraybackslash}X@{}}
        \textcolor{rqTwoP}{P1} & Define Applicability Boundaries & State the supported tasks, objects, and conditions, including cases in which the Skill does not apply. \\[2pt]
        \textcolor{rqTwoP}{P2} & Link References to Operations & At each operation that needs a reference, specify its location, when to consult it, and how to use it. \\[2pt]
        \textcolor{rqTwoP}{P3} & Specify Environment Requirements & List required tools, versions, and environment conditions, with readiness checks and setup steps. \\[2pt]
        \textcolor{rqTwoP}{P4} & Specify Access Requirements & State authentication methods and required permissions, and explain how to confirm resource access. \\[2pt]
      \end{tabularx}
    \end{tcolorbox}
  \par\nointerlineskip
    \begin{tcolorbox}[enhanced,frame hidden,sharp corners,boxrule=0pt,
      colback=white,colbacktitle=rqTwoA!14!white,coltitle=rqTwoA!65!black,
      borderline west={2.5pt}{0pt}{rqTwoA},
      left=5pt,right=1pt,top=2pt,bottom=2pt,boxsep=0pt,
      toptitle=3pt,bottomtitle=3pt,before skip=0pt,after skip=0pt,
      fonttitle=\footnotesize\sffamily,
      title={\textbf{ADAPTATION}\enspace {\scriptsize 3 practices}}]
      \begin{tabularx}{\linewidth}{@{}>{\raggedright\arraybackslash}p{0.035\linewidth}>{\scriptsize\raggedright\arraybackslash}p{0.31\linewidth}>{\raggedright\arraybackslash}X@{}}
        \textcolor{rqTwoA}{A1} & Explain Selection Criteria & Explain which task conditions justify each method or parameter choice. \\[2pt]
        \textcolor{rqTwoA}{A2} & Define Data Semantics & Define input and output meanings, including units, conventions, mappings, missing-value rules, and their basis. \\[2pt]
        \textcolor{rqTwoA}{A3} & Mark Replaceable Example Parameters & Mark values to replace and structures to retain when adapting examples or templates. \\[2pt]
      \end{tabularx}
    \end{tcolorbox}
  \par\nointerlineskip
    \begin{tcolorbox}[enhanced,frame hidden,sharp corners,boxrule=0pt,
      colback=white,colbacktitle=rqTwoI!14!white,coltitle=rqTwoI!65!black,
      borderline west={2.5pt}{0pt}{rqTwoI},
      left=5pt,right=1pt,top=2pt,bottom=2pt,boxsep=0pt,
      toptitle=3pt,bottomtitle=3pt,before skip=0pt,after skip=0pt,
      fonttitle=\footnotesize\sffamily,
      title={\textbf{IMPLEMENTATION}\enspace {\scriptsize 4 practices}}]
      \begin{tabularx}{\linewidth}{@{}>{\raggedright\arraybackslash}p{0.035\linewidth}>{\scriptsize\raggedright\arraybackslash}p{0.31\linewidth}>{\raggedright\arraybackslash}X@{}}
        \textcolor{rqTwoI}{I1} & Provide Referenced Resources & Include referenced scripts and resources or provide a valid way to obtain them; verify their locations. \\[2pt]
        \textcolor{rqTwoI}{I2} & Document Resource Invocation & Provide executable invocation instructions, required arguments, input formats, and expected outputs. \\[2pt]
        \textcolor{rqTwoI}{I3} & Specify Operation Dependencies & State necessary ordering and explain how intermediate files, data, or states pass to later operations. \\[2pt]
        \textcolor{rqTwoI}{I4} & Preserve Outcomes Across Alternatives & Define the results and constraints that alternative methods must preserve, including any additional steps required. \\[2pt]
      \end{tabularx}
    \end{tcolorbox}
  \par\nointerlineskip
    \begin{tcolorbox}[enhanced,frame hidden,sharp corners,boxrule=0pt,
      colback=white,colbacktitle=rqTwoV!14!white,coltitle=rqTwoV!65!black,
      borderline west={2.5pt}{0pt}{rqTwoV},
      left=5pt,right=1pt,top=2pt,bottom=2pt,boxsep=0pt,
      toptitle=3pt,bottomtitle=3pt,before skip=0pt,after skip=0pt,
      fonttitle=\footnotesize\sffamily,
      title={\textbf{VERIFICATION}\enspace {\scriptsize 3 practices}}]
      \begin{tabularx}{\linewidth}{@{}>{\raggedright\arraybackslash}p{0.035\linewidth}>{\scriptsize\raggedright\arraybackslash}p{0.31\linewidth}>{\raggedright\arraybackslash}X@{}}
        \textcolor{rqTwoV}{V1} & Define Acceptance Criteria & For each key requirement, identify what to check and the criteria for passing. \\[2pt]
        \textcolor{rqTwoV}{V2} & Validate Final Deliverables & Identify the actual deliverable and require checks of its contents and final state. \\[2pt]
        \textcolor{rqTwoV}{V3} & Specify Revalidation Triggers & Specify which changes invalidate earlier checks and which results must be checked again. \\[2pt]
      \end{tabularx}
    \end{tcolorbox}
  \par\nointerlineskip
    \begin{tcolorbox}[enhanced,frame hidden,sharp corners,boxrule=0pt,
      colback=white,colbacktitle=rqTwoC!14!white,coltitle=rqTwoC!65!black,
      borderline west={2.5pt}{0pt}{rqTwoC},
      left=5pt,right=1pt,top=2pt,bottom=2pt,boxsep=0pt,
      toptitle=3pt,bottomtitle=3pt,before skip=0pt,after skip=0pt,
      fonttitle=\footnotesize\sffamily,
      title={\textbf{CORRECTION}\enspace {\scriptsize 3 practices}}]
      \begin{tabularx}{\linewidth}{@{}>{\raggedright\arraybackslash}p{0.035\linewidth}>{\scriptsize\raggedright\arraybackslash}p{0.31\linewidth}>{\raggedright\arraybackslash}X@{}}
        \textcolor{rqTwoC}{C1} & Explain Diagnosis and Recovery & Explain how to diagnose relevant failures and which corrective actions to take. \\[2pt]
        \textcolor{rqTwoC}{C2} & Define Retry Boundaries & State what must change before retrying and when to stop or switch methods. \\[2pt]
        \textcolor{rqTwoC}{C3} & Preserve Justified Acceptance Criteria & Preserve justified acceptance requirements during repair; revise them only with evidence that they are incorrect or inapplicable. \\[2pt]
      \end{tabularx}
    \end{tcolorbox}
  \par\smallskip
\end{table}

\subsubsection{Utility-Grounded Authoring Practices}

Beyond task relevance and operation matching, downstream utility also relates to how Skill content specifies conditions of use, execution procedures, and required outcomes. We organize the authoring requirements identified through content and execution analysis into the 17 practices in Table~\ref{tab:rq2-best-practices}. The framework groups practices into five stages by their primary use. Each applies to the relevant operations or resources.

Preparation and Adaptation address the conditions under which Skill support can be used for the current task. Preparation specifies applicability boundaries, reference use, and environment and access prerequisites (P1--P4), helping the agent assess applicability and prepare for use. Adaptation connects general guidance to the task by explaining method and parameter choices, input and output semantics, and the parts of examples that require replacement (A1--A3). These requirements explain how relevant capabilities can be used in the current task, addressing execution gaps left by unclear prerequisites or insufficient task adaptation.

Implementation concerns whether Skill content translates its stated capabilities into concrete operations. Beyond naming scripts or describing tools, content needs to provide accessible resources and explicit invocation instructions, including required parameters, input formats, and expected outputs (I1--I2). For procedures involving multiple steps, it should explain how intermediate data or states pass between operations and which outcomes and constraints alternative methods must preserve (I3--I4). These practices address the executability of operation support, guiding resource invocation, step coordination, and method substitution while reducing what the agent must supply to turn capability descriptions into concrete actions.

Verification and Correction address whether execution results meet task requirements and how to proceed when execution is blocked. Verification specifies acceptance criteria, the actual deliverable, and conditions for revalidation after changes (V1--V3), so that checks cover the final delivery state. Correction explains diagnosis and recovery, defines when to retry or switch methods, and preserves justified acceptance requirements during repair (C1--C3). Together, these practices connect operation completion, result checking, and failure handling to the delivery of results that meet task requirements, enabling Skill support to contribute to task gains.

\begin{findingbox}{Finding 2c}
\textbf{Skill authoring for utility links executable procedures to recovery and checks on the final deliverable.}
The identified practices emphasize rechecking results after changes and preserving justified acceptance criteria during repair.
\end{findingbox}

\subsection{RQ3: Effects of Multi-Skill Organization on Utility}
\label{subsec:rq3-results}
\suppressfloats[t]

\begin{table}[t]
  \caption{Pass rates (\%) under alternative organizations of the same Skill set.}
  \label{tab:rq3-pass}
  \centering\small
  \setlength{\tabcolsep}{3pt}
  \begin{tabular}{@{}lccccc@{}}
    \toprule
    Configuration & No-Skill & Flat & Sequence & \shortstack{Stage\\Plan} & \shortstack{Dependency\\DAG} \\
    \midrule
    GPT + Codex & 44.72 & 52.03 & 52.85 & 55.28 & \textbf{58.54} \\
    DeepSeek + OpenClaw & 34.96 & 39.84 & 47.97 & \textbf{51.22} & \textbf{51.22} \\
    Qwen + OpenCode & 11.38 & 21.14 & 21.95 & 25.20 & \textbf{27.64} \\
    \bottomrule
  \end{tabular}
\end{table}

The downstream utility of the same Skills varies with their organization, with schemes specifying stage objectives or artifact dependencies outperforming order alone. As Table~\ref{tab:rq3-pass} shows, Sequence, Stage Plan, and Dependency DAG all achieve higher overall pass rates than Flat in all three configurations, while Stage Plan and DAG also outperform Sequence. DAG improves over Flat by 6.50, 11.38, and 6.50 percentage points under GPT + Codex, DeepSeek + OpenClaw, and Qwen + OpenCode, respectively, achieving the highest or tied-highest overall pass rate. All conditions provide the same Skills and supporting resources. Organizing their use thus yields additional gains, whose magnitude varies across organization schemes.

\begin{figure*}[t]
  \centering
  \includegraphics[width=\linewidth]{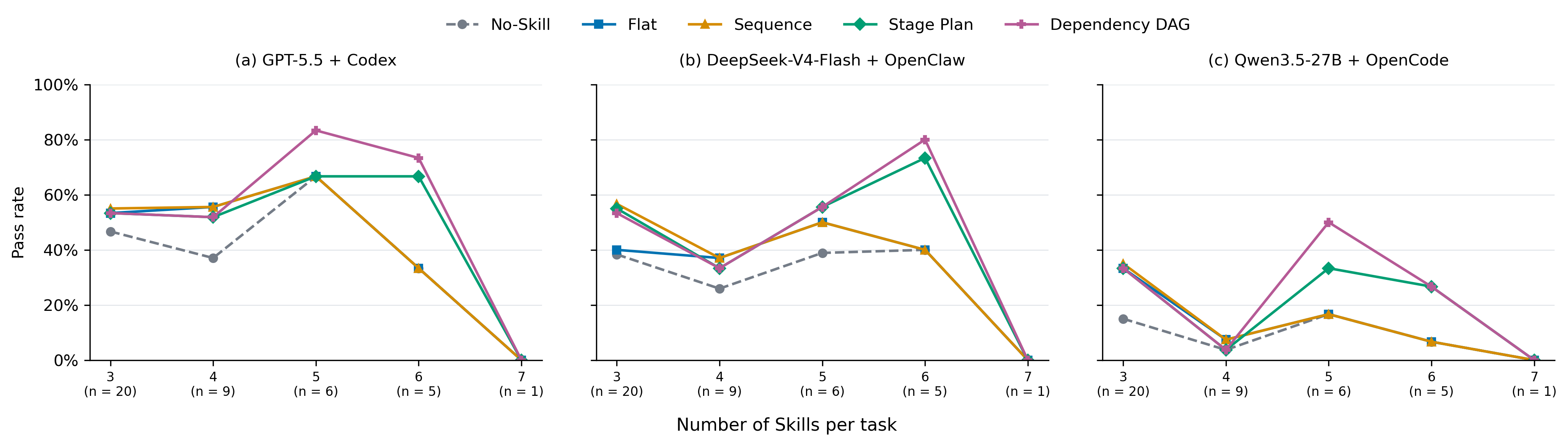}
  \caption{Pass rates by the number of supplied Skills.}
  \Description{Three line charts compare No-Skill, Flat, Sequence, Stage Plan, and Dependency DAG across groups with three to seven supplied Skills. Stage Plan and DAG have identical overall pass rates in the four-Skill group in all configurations and in the three-Skill group for GPT and Qwen, with a small difference for DeepSeek. Their gains over Flat are larger in the combined five- and six-Skill group than in the combined three- and four-Skill group in each configuration.}
  \label{fig:rq3-skill-count}
\end{figure*}

DAG's additional gains over Stage Plan occur mainly among tasks supplied with five or six Skills. The two schemes have identical overall pass rates in the three-Skill group for GPT and Qwen and differ by only 1.67 percentage points for DeepSeek (Figure~\ref{fig:rq3-skill-count}). In the four-Skill group, their pass rates are identical across all three configurations. In contrast, across the 11 tasks supplied with five or six Skills, DAG improves over Stage Plan by 12.12, 3.03, and 9.09 percentage points in the three configurations, respectively. In this group, Stage Plan improves over Flat by 15.15--18.18 percentage points and DAG by 21.21--27.27 points, both exceeding their gains in the combined three- and four-Skill group under the corresponding configuration. Stage Plan benefits this task group, while the DAG scheme yields further gains.

Execution trajectories show that intermediate artifacts specified in organization instructions serve both to connect operations and to support checks and revisions. For example, \texttt{jpg-\allowbreak ocr-\allowbreak stat} requires extracting dates and total amounts from receipts and generating a workbook. Stage Plan assigns OCR and field extraction to one stage and workbook generation to another. DAG instead requires saving the OCR text produced using \texttt{image-\allowbreak ocr}, then extracting and saving structured fields for subsequent workbook generation using \texttt{xlsx}. GPT + Codex passes 0/3 runs under Flat, Sequence, and Stage Plan, but 3/3 under DAG. In one successful DAG run, the agent inspects the extracted fields, reads the OCR text for the corresponding receipts, then reruns field extraction, updates the field file, and rebuilds the workbook. The OCR text preserves the source of the fields, while the field file connects extraction to workbook generation. The agent can revisit these intermediate results and carry revisions through to the final deliverable. These artifacts support both cross-Skill result transfer and correction before delivery.

\begin{findingbox}{Finding 3}
\textbf{Organizing the same Skills around stage objectives or artifact dependencies yields greater utility than specifying use order alone.}
Intermediate artifacts let agents check results and carry corrections through to the final deliverable.
\end{findingbox}

\section{Discussion}
\label{sec:discussion}


\subsection{Addressing the Study Challenges}
\label{subsec:addressing-challenges}

Our study addresses both challenges through comparisons of real-world Skills
and analysis of their actual use. For Challenge 1, we compare relevant
published alternatives on the same tasks under common configurations, retain
their original instructions and resources, and examine the same Skills across
configurations and organization strategies. These comparisons ground the 17
authoring practices in observed content and execution differences. For
example, the unit-harmonization comparison reveals different support for
identifying units and cleaning records despite both candidates matching the
task topic. Comparing the support actually available helps identify concrete
requirements that a topic-level assessment would miss. For
Challenge 2, we jointly examine Skill content, execution traces, and final
artifacts to distinguish guidance agents follow, skip, or adapt from
operations they supply themselves. LLM-assisted analysis and author review
connect these observations to task outcomes, providing evidence for
interpreting utility differences without treating the presence of guidance as
proof of its use. Reviewing alternative explanations and cases with smaller
or no outcome differences also helps clarify the scope of the inferred
practices. These comparisons support empirical interpretations, while the
effects of individual content requirements need controlled validation.

\subsection{Implications for Skill Development and Research}
\label{subsec:skill-implications}

\noindent\textbf{\textit{Skill Development and Reuse.}} RQ1 and RQ2 motivate
making operation support, prerequisites, and evaluated configurations explicit
when publishing Skills. This information can guide candidate selection, while
the authoring framework supports adaptation that preserves task requirements
and checks final artifacts. Building on prior work on Skill quality and
adaptation~\cite{hong2026anatomy,wu2026adaptation}, the practices can inform
review tools and revision suggestions. Representative tasks can also support
regression checks when Skill content or execution configurations change.
For authors, this means making clear which methods can change, which task
requirements remain binding, and which checks must be repeated after a
revision. For adopters, candidate descriptions could distinguish directly
usable support from operations requiring additional logic. Recording the
tasks and configurations used for validation would further help users judge
whether reported gains apply to their setting and decide what to retest
before deployment.

\noindent\textbf{\textit{Harness Design and Automation.}} Our findings motivate
harness support that connects candidate assessment, execution, and multi-Skill
coordination. Tools could identify unsupported operations, check prerequisites,
and track which guidance agents use. RQ3 further suggests deriving stage
objectives and artifact dependencies, retaining intermediate results, and
flagging downstream operations or checks affected by changes. Automating
these capabilities would connect planning with recovery and revision, while
requiring evaluation of the correctness and cost of automatically generated
organization instructions.
The organization results suggest that such tools need to represent both the
results passed between Skills and the conditions for checking them. When an
upstream artifact changes, its consumers may need to rerun and earlier checks
may become invalid. A harness could track these relationships and retain
source artifacts for inspection. Our manually constructed organization
schemes provide a starting point for studying automatic inference of these
dependencies and their maintenance during execution.

\noindent\textbf{\textit{Evaluation and Generalization.}} The authoring practices
provide concrete variables for controlled experiments on individual content
changes and their interactions. Combining pass rates with evidence of guidance
use can help explain their effects, following the process perspective of
Skill-Use~\cite{han2026skill}. Further evaluation can test transfer across
tasks, configurations, and Skill combinations, including selection and
interference in larger libraries~\cite{huang2026consumption}. Execution time,
token use, and maintenance effort can help assess whether the gains justify
adoption costs.
Experiments can vary recovery guidance, acceptance criteria, and revalidation
triggers separately, then test their combinations to identify interactions.
Recording whether agents read and act on the revised instructions can help
distinguish limited adoption from ineffective guidance. Independent task sets
are also needed to evaluate the stability of operation-support reranking.
Together, these evaluations can clarify when richer Skill content or more
explicit organization is worth its additional execution and maintenance cost.

\section{Threats to Validity}
\label{sec:threats}

\noindent\textbf{\textit{Construct Validity.}} We use the original task verifiers to compare pass rates under consistent acceptance criteria, and examine execution trajectories and final artifacts to assess how Skill content is used. This measure captures downstream utility in terms of task completion, leaving partial completion and execution cost outside its scope. Atomic operation-based reranking uses common operation representations, support labels, and scoring rules, with utility evaluated through candidates' actual execution outcomes. LLM-based operation extraction and support judgments use source evidence but may still introduce annotation errors that affect candidate rankings. The scores assigned to support labels are heuristic choices. Different label scores or operation granularities could change candidate rankings and the resulting selection gains.

\noindent\textbf{\textit{Internal Validity.}} Each run starts from a fresh environment, and we compare conditions on the same task under the same configuration, which reduces interference from residual state and configuration differences. Skill conditions provide both content and an explicit request to consult and use it, whereas No-Skill provides neither. Gains relative to No-Skill therefore reflect this combined setting. RQ2 retains candidates' original content and resources and compares content with execution evidence to identify authoring features associated with utility. Because candidate attributes may vary together, these insights concern associations observed during reuse. RQ3 retains the same available Skill set, with organization instructions constructed solely from task descriptions and the corresponding Skills and reviewed by another author. Skill-use order, stage objectives, and artifact handoffs still involve author choices, and we therefore interpret gains at the level of complete organization schemes.

\noindent\textbf{\textit{External Validity.}} The study covers 87 tasks across eight domains, three models, and three harnesses, and draws candidate Skills from snapshots of two public platforms to compare utility across tasks and configurations. Corpus filters focus on English-language Skills with complete content and some evidence of adoption. RQ2 and RQ3 use 23 and 41 tasks, respectively, under the selected configuration for each model. RQ2 candidates are restricted to the top five Skills retrieved for each task. Generalization to other languages, newly published or less widely adopted Skills, open-ended workflows, and other configurations requires further study.

\noindent\textbf{\textit{Conclusion Validity.}} We repeat each condition three times, average outcomes within each task, and then average equally across tasks to reduce the influence of individual runs on aggregate comparisons. With only three repetitions per task and condition, task-level pass rates have limited resolution and remain sensitive to execution randomness~\cite{yao2024taubench}. RQ2 compares gains before and after reranking using the same candidate set and existing execution records, focusing the comparison on changes in selection order. Because the reranking method was motivated by observations on the current task set, the stability of its gains requires evaluation on independent tasks. RQ3 examines organization gains by supplied Skill count. The five- and six-Skill groups contain six and five tasks, respectively, so group differences may also reflect task composition.

\section{Conclusion}
\label{sec:conclusion}

We study the downstream utility of Agent Skills on 87 SkillsBench tasks, combining controlled execution with comparisons of Skill content and analysis of execution trajectories. 
The utility of the same Skills varies across model--harness configurations, and relevance rankings do not fully reflect candidate utility. 
Reranking by task-operation support improves first-choice pass rates across the three evaluated configurations. 
With the Skill set unchanged, schemes specifying stage objectives or artifact dependencies achieve higher overall pass rates than specifying order alone. Content and trajectory comparisons provide case evidence for authoring practices concerning deliverable verification, revalidation after changes, and dependency recovery, organized within a five-stage Skill authoring framework. 
These findings link Skill content, actual use, and task outcomes, informing decisions about Skill authoring, selection, and organization.

\section*{Data Availability}
All materials supporting this study, including experimental data, execution records, analysis scripts, and paper sources, 
are available at \url{https://anonymous.4open.science/r/skill_utility-D2C6}.

\bibliographystyle{ACM-Reference-Format-citation-order}
\bibliography{references}

@misc{anthropic2025skillspec,
  author       = {{Anthropic and the Agent Skills Community}},
  title        = {Agent Skills Specification},
  year         = {2025},
  howpublished = {\url{https://agentskills.io/specification}},
  note         = {Accessed 2026-09-17}
}

@article{han2026skill,
  title={Skill-Use: Can LLMs Actually Use Skills in Agentic Harnesses?},
  author={Han, Jinyi and Xu, Yuanjian and Liao, Ying and Wang, Xinyi and Jiang, Zishang and Di, Zixiang and Lu, Fanyang and Hu, Zhichao and Xiao, Yanghua},
  journal={arXiv preprint arXiv:2608.04828},
  year={2026}
}

@article{hong2026anatomy,
  title={From Anatomy to Smells: An Empirical Study of SKILL. md in Agent Skills},
  author={Hong, David Boram and Imani, Aaron and Ahmed, Iftekhar},
  journal={arXiv preprint arXiv:2607.01456},
  year={2026}
}

@article{chen2026skvm,
  title={Skvm: Revisiting language vm for skills across heterogenous llms and harnesses},
  author={Chen, Le and Feng, Erhu and Xia, Yubin and Chen, Haibo},
  journal={arXiv preprint arXiv:2604.03088},
  year={2026}
}

@article{liang2026ssl,
  title={From skill text to skill structure: The scheduling-structural-logical representation for agent skills},
  author={Liang, Qiliang and Wang, Hansi and Liang, Zhong and Liu, Yang},
  journal={arXiv preprint arXiv:2604.24026},
  year={2026}
}

@article{ling2026skills,
  title={Agent skills: A data-driven analysis of claude skills for extending large language model functionality},
  author={Ling, George and Zhong, Shanshan and Huang, Richard},
  journal={arXiv preprint arXiv:2602.08004},
  year={2026}
}

@article{gao2026skillreducer,
  title={Skillreducer: Optimizing llm agent skills for token efficiency},
  author={Gao, Yudong and Li, Zongjie and Yuan, Yuanyuan and Ji, Zimo and Ma, Pingchuan and Wang, Shuai},
  journal={arXiv preprint arXiv:2603.29919},
  year={2026}
}

@article{huang2026generated,
  title={Do LLM-Generated Skills Make Better AI Data Scientists? A Component Ablation Across Data-Science Workflows},
  author={Huang, Wei-Jung},
  journal={arXiv preprint arXiv:2607.07504},
  year={2026}
}

@article{huang2026consumption,
  title={From raw experience to skill consumption: A systematic study of model-generated agent skills},
  author={Huang, Zisu and Xu, Jingwen and Yang, Yifan and Gong, Ziyang and Yang, Qihao and Tian, Muzhao and Wang, Xiaohua and Lv, Changze and Gao, Xuemei and Dai, Qi and others},
  journal={arXiv preprint arXiv:2605.23899},
  year={2026}
}

@article{wu2026adaptation,
  title={An Empirical Study of Downstream Adaptation for Agent Skills},
  author={Wu, Xinjian and Gong, Jingzhi and Jahangirova, Gunel and Chen, Zhenpeng and Zhang, Jie M},
  journal={arXiv preprint arXiv:2607.03238},
  year={2026}
}

@article{li2026skillsbench,
  title={SkillsBench: Benchmarking how well agent skills work across diverse tasks},
  author={Li, Xiangyi and Liu, Yimin and Chen, Wenbo and You, Bingran and Di, Zonglin and He, Yifeng and Zheng, Shenghan and Choe, Kyoung Whan and Sun, Jiankai and Wang, Shuyi and others},
  journal={arXiv preprint arXiv:2602.12670},
  year={2026}
}

@article{han2026sweskills,
  title={Do Agent Skills Actually Help in Real-World Software Engineering},
  author={Han, Tingxu and Zhang, Yi and Song, Wei and Fang, Chunrong and Chen, Zhenyu and Sun, Youcheng},
  journal={arXiv preprint arXiv:2603.15401},
  year={2026}
}

@inproceedings{yao2023react,
title={ReAct: Synergizing Reasoning and Acting in Language Models},
author={Shunyu Yao and Jeffrey Zhao and Dian Yu and Nan Du and Izhak Shafran and Karthik R Narasimhan and Yuan Cao},
booktitle={The Eleventh International Conference on Learning Representations },
year={2023},
url={https://openreview.net/forum?id=WE_vluYUL-X}
}

@inproceedings{yang2024sweagent,
 author = {Yang, John and Jimenez, Carlos and Wettig, Alexander and Lieret, Kilian and Yao, Shunyu and Narasimhan, Karthik and Press, Ofir},
 booktitle = {Advances in Neural Information Processing Systems},
 doi = {10.52202/079017-1601},
 editor = {A. Globerson and L. Mackey and D. Belgrave and A. Fan and U. Paquet and J. Tomczak and C. Zhang},
 pages = {50528--50652},
 publisher = {Curran Associates, Inc.},
 title = {SWE-agent: Agent-Computer Interfaces Enable Automated Software Engineering},
 url = {https://proceedings.neurips.cc/paper_files/paper/2024/file/5a7c947568c1b1328ccc5230172e1e7c-Paper-Conference.pdf},
 volume = {37},
 year = {2024}
}

@inproceedings{wang2025openhands,
  title={Openhands: An open platform for ai software developers as generalist agents},
  author={Wang, Xingyao and Li, Boxuan and Song, Yufan and Xu, Frank F and Tang, Xiangru and Zhuge, Mingchen and Pan, Jiayi and Song, Yueqi and Li, Bowen and Singh, Jaskirat and others},
  booktitle={International Conference on Learning Representations},
  volume={2025},
  pages={65882--65919},
  year={2025}
}

@article{zheng2026skillrouter,
  title={Skillrouter: Skill routing for llm agents at scale},
  author={Zheng, YanZhao and Zhang, ZhenTao and Ma, Chao and Yu, YuanQiang and Zhu, JiHuai and Wu, Yong and Xu, Tianze and Dong, Baohua and Zhu, Hangcheng and Huang, Ruohui and others},
  journal={arXiv preprint arXiv:2603.22455},
  year={2026}
}

@article{cho2026skillret,
  title={SkillRet: A large-scale benchmark for skill retrieval in LLM agents},
  author={Kang, Ryangkyung and Cho, Hongcheol and Kim, Youngeun},
  journal={arXiv preprint arXiv:2605.05726},
  year={2026}
}

@article{su2026sra,
  title={Skill retrieval augmentation for agentic ai},
  author={Su, Weihang and Long, Jianming and Ai, Qingyao and He, Qiaozhi and Tang, Yichen and Wang, Changyue and Tu, Yiteng and Wang, Yingbo and Liu, Yiqun},
  journal={arXiv preprint arXiv:2604.24594},
  year={2026}
}

@article{xia2026grasp,
  title={Grasp: Graph-structured skill compositions for llm agents},
  author={Xia, Tianle and Hu, Lingxiang and Sun, Yiding and Xu, Ming and Xu, Lan and Wang, Siying and Xu, Wei and Jiang, Jie},
  journal={arXiv preprint arXiv:2604.17870},
  year={2026}
}

@article{gong2026skillmoo,
  title={Skillmoo: Multi-objective optimization of agent skills for software engineering},
  author={Gong, Jingzhi and Gu, Ruizhen and Fei, Zhiwei and Cao, Yazhuo and Twist, Lukas and Geiger, Alina and Han, Shuo and Sobania, Dominik and Sarro, Federica and Zhang, Jie M},
  journal={arXiv preprint arXiv:2604.09297},
  year={2026}
}

@article{ouyang2026skcc,
  title={Skcc: Portable and secure skill compilation for cross-framework llm agents},
  author={Ouyang, Yipeng and Xiao, Yi and Gu, Yuhao and Zhang, Xianwei},
  journal={arXiv preprint arXiv:2605.03353},
  year={2026}
}

@article{liu2026skillswild,
  title={Agent skills in the wild: An empirical study of security vulnerabilities at scale},
  author={Liu, Yi and Wang, Weizhe and Feng, Ruitao and Zhang, Yao and Xu, Guangquan and Deng, Gelei and Li, Yuekang and Zhang, Leo},
  journal={arXiv preprint arXiv:2601.10338},
  year={2026}
}

@article{li2026graphskills,
  title={Graph-of-Skills: Dependency-Aware Structural Retrieval for Massive Agent Skills},
  author={Liu, Dawei and Li, Zongxia and Du, Hongyang and Wu, Xiyang and Gui, Shihang and Kuang, Yongbei and Sun, Lichao},
  journal={arXiv preprint arXiv:2604.05333},
  year={2026}
}

@misc{bai2026skilldag,
  title={SkillDAG: Self-Evolving Typed Skill Graphs for LLM Skill Selection at Scale},
  author={Bai, Tong and Wan, Zhenglin and Zhou, Pengfei and Yu, Xingrui and You, Yang and Tsang, Ivor W},
  journal={arXiv preprint arXiv:2606.03056},
  year={2026}
}

@article{wang2024voyager,
  title={An open-ended embodied agent with large language models},
  author={Wang, G and Xie, Y and Jiang, Y and Mandlekar, A and Xiao, C and Zhu, Y and Fan, L and Anandkumar, AV},
  journal={arXiv preprint arxiv:2305.16291},
  year={2023}
}

@inproceedings{zhao2024expel,
  title={Expel: Llm agents are experiential learners},
  author={Zhao, Andrew and Huang, Daniel and Xu, Quentin and Lin, Matthieu and Liu, Yong-Jin and Huang, Gao},
  booktitle={Proceedings of the AAAI Conference on Artificial Intelligence},
  volume={38},
  number={17},
  pages={19632--19642},
  year={2024}
}

@inproceedings{wang2025awm,
  title={Agent Workflow Memory},
  author={Zora Zhiruo Wang and Jiayuan Mao and Daniel Fried and Graham Neubig},
  booktitle={Forty-second International Conference on Machine Learning},
  year={2025},
  url={https://openreview.net/forum?id=NTAhi2JEEE}
}

@inproceedings{wang2025programmatic,
  title={Inducing programmatic skills for agentic tasks},
  author={Wang, Zora Zhiruo and Gandhi, Apurva and Neubig, Graham and Fried, Daniel},
  journal={arXiv preprint arXiv:2504.06821},
  year={2025}
}

@article{liang2026skillnet,
  title={Skillnet: Create, evaluate, and connect ai skills},
  author={Liang, Yuan and Zhong, Ruobin and Xu, Haoming and Jiang, Chen and Zhong, Yi and Fang, Runnan and Gu, Jia-Chen and Deng, Shumin and Yao, Yunzhi and Wang, Mengru and others},
  journal={arXiv preprint arXiv:2603.04448},
  year={2026}
}

@article{robertson2009bm25,
  title={The probabilistic relevance framework: BM25 and beyond},
  author={Robertson, Stephen and Zaragoza, Hugo},
  journal={Foundations and trends{\textregistered} in information retrieval},
  volume={4},
  number={1-2},
  pages={1--174},
  year={2009},
  publisher={Emerald Publishing Limited}
}

@inproceedings{shi2025toolret,
  title={Retrieval models aren’t tool-savvy: Benchmarking tool retrieval for large language models},
  author={Shi, Zhengliang and Wang, Yuhan and Yan, Lingyong and Ren, Pengjie and Wang, Shuaiqiang and Yin, Dawei and Ren, Zhaochun},
  booktitle={Findings of the Association for Computational Linguistics: ACL 2025},
  pages={24497--24524},
  year={2025}
}

@inproceedings{lu2026tooldocs,
  title={Tools are under-documented: Simple document expansion boosts tool retrieval},
  author={Lu, Xuan and Huang, Haohang and Meng, Rui and Jin, Yaohui and Zeng, Wenjun and Shen, Xiaoyu},
  booktitle={International Conference on Learning Representations},
  volume={2026},
  pages={141631--141653},
  year={2026}
}

@inproceedings{kim2024llmcompiler,
title={An {LLM} Compiler for Parallel Function Calling},
author={Sehoon Kim and Suhong Moon and Ryan Tabrizi and Nicholas Lee and Michael W. Mahoney and Kurt Keutzer and Amir Gholami},
booktitle={Forty-first International Conference on Machine Learning},
year={2024},
url={https://openreview.net/forum?id=uQ2FUoFjnF}
}

@article{li2026agentskillos,
  title={Organizing, orchestrating, and benchmarking agent skills at ecosystem scale},
  author={Li, Hao and Mu, Chunjiang and Chen, Jianhao and Ren, Siyue and Cui, Zhiyao and Zhang, Yiqun and Bai, Lei and Hu, Shuyue},
  journal={arXiv preprint arXiv:2603.02176},
  year={2026}
}

@misc{openclawDocs,
  author       = {{OpenClaw}},
  title        = {{OpenClaw} Documentation},
  howpublished = {\url{https://docs.openclaw.ai/}},
  note         = {Accessed 2026-09-17}
}

@misc{benchflow2026skillsbenchtasks,
  author       = {{SkillsBench Contributors}},
  title        = {{SkillsBench} Task Directory},
  year         = {2026},
  howpublished = {\url{https://github.com/benchflow-ai/skillsbench/tree/main/tasks}},
  note         = {Accessed 2026-09-20}
}

@article{wang2026assay,
  title={Not All Skills Help: Measuring and Repairing Agent Knowledge},
  author={Wang, Yixuan and Zhou, Yiyang and Liang, Yiming and Zhang, Congyu and Liu, Fuxiao and Zhou, Jiawei and Yao, Huaxiu},
  journal={arXiv preprint arXiv:2606.15390},
  year={2026}
}

@inproceedings{cruzes2011thematic,
  title={Recommended steps for thematic synthesis in software engineering},
  author={Cruzes, Daniela S and Dyba, Tore},
  booktitle={2011 international symposium on empirical software engineering and measurement},
  pages={275--284},
  year={2011},
  organization={IEEE}
}

@article{runeson2009casestudy,
  title={Guidelines for conducting and reporting case study research in software engineering},
  author={Runeson, Per and H{\"o}st, Martin},
  journal={Empirical software engineering},
  volume={14},
  number={2},
  pages={131--164},
  year={2009},
  publisher={Springer}
}

@inproceedings{sun2023rankgpt,
  title={Is ChatGPT good at search? investigating large language models as re-ranking agents},
  author={Sun, Weiwei and Yan, Lingyong and Ma, Xinyu and Wang, Shuaiqiang and Ren, Pengjie and Chen, Zhumin and Yin, Dawei and Ren, Zhaochun},
  booktitle={Proceedings of the 2023 conference on empirical methods in natural language processing},
  pages={14918--14937},
  year={2023}
}

@misc{claudepluginsRegistry,
  title = {{Claude Code Plugins \& Agent Skills}---Community Registry},
  key = {claude-plugins.dev},
  url = {https://claude-plugins.dev/},
  note = {Accessed 2026-09-26}
}

@misc{clawhubRegistry,
  author = {{OpenClaw}},
  title = {{ClawHub}---Public Skill Registry},
  url = {https://github.com/openclaw/clawhub},
  note = {Accessed 2026-09-26}
}

@misc{openai2026gpt55,
  author = {{OpenAI}},
  title = {{GPT-5.5} Model},
  year = {2026},
  url = {https://developers.openai.com/api/docs/models/gpt-5.5},
  note = {Accessed 2026-09-24}
}

@misc{deepseek2026v4flash,
  author = {{DeepSeek}},
  title = {{DeepSeek-V4-Flash} Model Card},
  year = {2026},
  url = {https://huggingface.co/deepseek-ai/DeepSeek-V4-Flash},
  note = {Accessed 2026-09-24}
}

@misc{qwen2026qwen35_27b,
  author = {{Qwen Team}},
  title = {{Qwen3.5-27B} Model Card},
  year = {2026},
  url = {https://huggingface.co/Qwen/Qwen3.5-27B},
  note = {Accessed 2026-09-24}
}

@misc{openai2026codexcli,
  author = {{OpenAI}},
  title = {{Codex CLI}},
  year = {2026},
  url = {https://github.com/openai/codex},
  note = {Accessed 2026-09-24}
}

@misc{opencode2026docs,
  author = {{OpenCode}},
  title = {{OpenCode} Documentation},
  year = {2026},
  url = {https://opencode.ai/docs/},
  note = {Accessed 2026-09-24}
}

@inproceedings{wang2023plansolve,
  title={Plan-and-solve prompting: Improving zero-shot chain-of-thought reasoning by large language models},
  author={Wang, Lei and Xu, Wanyu and Lan, Yihuai and Hu, Zhiqiang and Lan, Yunshi and Lee, Roy Ka-Wei and Lim, Ee-Peng},
  booktitle={Proceedings of the 61st annual meeting of the association for computational linguistics (volume 1: Long papers)},
  pages={2609--2634},
  year={2023}
}

@inproceedings{ahmed2025annotation,
  author={Ahmed, Toufique and Devanbu, Premkumar and Treude, Christoph and Pradel, Michael},
  booktitle={2025 IEEE/ACM 22nd International Conference on Mining Software Repositories (MSR)}, 
  title={Can LLMs Replace Manual Annotation of Software Engineering Artifacts?}, 
  year={2025},
  volume={},
  number={},
  pages={526-538},
  doi={10.1109/MSR66628.2025.00086}
}

@article{wang2025llmjudge,
  title={Can llms replace human evaluators? an empirical study of llm-as-a-judge in software engineering},
  author={Wang, Ruiqi and Guo, Jiyu and Gao, Cuiyun and Fan, Guodong and Chong, Chun Yong and Xia, Xin},
  journal={Proceedings of the ACM on Software Engineering},
  volume={2},
  number={ISSTA},
  pages={1955--1977},
  year={2025},
  publisher={ACM New York, NY, USA}
}

@inproceedings{schick2023toolformer,
  title={Toolformer: Language models can teach themselves to use tools},
  author={Schick, Timo and Dwivedi-Yu, Jane and Dess{\`\i}, Roberto and Raileanu, Roberta and Lomeli, Maria and Hambro, Eric and Zettlemoyer, Luke and Cancedda, Nicola and Scialom, Thomas},
  booktitle={Advances in Neural Information Processing Systems},
  volume={36},
  pages={68539--68551},
  year={2023}
}

@inproceedings{qin2024toolllm,
  title={Toolllm: Facilitating large language models to master 16000+ real-world apis},
  author={Qin, Yujia and Liang, Shihao and Ye, Yining and Zhu, Kunlun and Yan, Lan and Lu, Yaxi and Lin, Yankai and Cong, Xin and Tang, Xiangru and Qian, Bill and others},
  booktitle={International Conference on Learning Representations},
  volume={2024},
  pages={9695--9717},
  year={2024}
}

@inproceedings{shinn2023reflexion,
  title={Reflexion: Language agents with verbal reinforcement learning},
  author={Shinn, Noah and Cassano, Federico and Gopinath, Ashwin and Narasimhan, Karthik and Yao, Shunyu},
  booktitle={Advances in Neural Information Processing Systems},
  volume={36},
  pages={8634--8652},
  year={2023}
}

@article{xia2026skillrl,
  title={Skillrl: Evolving agents via recursive skill-augmented reinforcement learning},
  author={Xia, Peng and Chen, Jianwen and Wang, Hanyang and Liu, Jiaqi and Zeng, Kaide and Wang, Yu and Han, Siwei and Zhou, Yiyang and Zhao, Xujiang and Chen, Haifeng and others},
  journal={arXiv preprint arXiv:2602.08234},
  year={2026}
}

@inproceedings{liu2024controlllm,
  title={Controlllm: Augment language models with tools by searching on graphs},
  author={Liu, Zhaoyang and Lai, Zeqiang and Gao, Zhangwei and Cui, Erfei and Li, Ziheng and Zhu, Xizhou and Lu, Lewei and Chen, Qifeng and Qiao, Yu and Dai, Jifeng and others},
  booktitle={European conference on computer vision},
  pages={89--105},
  year={2024},
  organization={Springer}
}

@inproceedings{sun2023adaplanner,
  title={Adaplanner: Adaptive planning from feedback with language models},
  author={Sun, Haotian and Zhuang, Yuchen and Kong, Lingkai and Dai, Bo and Zhang, Chao},
  booktitle={Advances in Neural Information Processing Systems},
  volume={36},
  pages={58202--58245},
  year={2023}
}

@inproceedings{singh2023progprompt,
  author={Singh, Ishika and Blukis, Valts and Mousavian, Arsalan and Goyal, Ankit and Xu, Danfei and Tremblay, Jonathan and Fox, Dieter and Thomason, Jesse and Garg, Animesh},
  booktitle={2023 IEEE International Conference on Robotics and Automation (ICRA)}, 
  title={ProgPrompt: Generating Situated Robot Task Plans using Large Language Models}, 
  year={2023},
  volume={},
  number={},
  pages={11523-11530},
  doi={10.1109/ICRA48891.2023.10161317}
}

@article{yao2024taubench,
  title={{$\tau$-bench}: A Benchmark for Tool-Agent-User Interaction in Real-World Domains},
  author={Yao, Shunyu and Shinn, Noah and Razavi, Pedram and Narasimhan, Karthik},
  journal={arXiv preprint arXiv:2406.12045},
  year={2024}
}

@article{zeng2026groupskills,
  title={Group of skills: Group-structured skill retrieval for agent skill libraries},
  author={Zeng, Kun and Huo, Yu and Zhang, Siyu and Ye, Zi and Zhuo, Yuecheng and Liu, Haoyue and Lu, Yuquan and Wen, Junhao and Tang, Xiaoying},
  journal={arXiv preprint arXiv:2605.06978},
  year={2026}
}
\end{document}